\documentclass[runningheads,a4paper]{llncs}

\usepackage{amssymb}
\usepackage{graphicx}
\usepackage{float}

\usepackage{url}
\urldef{\mailsa}\path|{alfred.hofmann, ursula.barth, ingrid.haas, frank.holzwarth,|
\urldef{\mailsb}\path|anna.kramer, leonie.kunz, christine.reiss, nicole.sator,|
\urldef{\mailsc}\path|erika.siebert-cole, peter.strasser, lncs}@springer.com|    
\newcommand{\keywords}[1]{\par\addvspace\baselineskip
\noindent\keywordname\enspace\ignorespaces#1}

\floatstyle{ruled}
\newfloat{algorithm}{tbp}{loa}
\providecommand{\algorithmname}{Algorithm}

\floatname{algorithm}{\protect\algorithmname}
\usepackage{amsmath,graphicx}
\usepackage{epsfig}
\usepackage{graphics} 
\usepackage{epsfig} 
\usepackage{amsmath}
\usepackage{url}
\usepackage{tablefootnote}
\usepackage{subfig}
\usepackage{upgreek}

\begin{document}

\mainmatter  

\title{Joint Segmentation and Uncertainty Visualization of Retinal Layers in Optical Coherence Tomography Images using Bayesian Deep Learning}

\titlerunning{Segmentation of retinal layers in optical coherence tomography images using Bayesian deep learning}

%
%
%


\author{Suman Sedai,  Bhavna Antony, Dwarikanath Mahapatra  and Rahil Garnavi}
\institute{IBM Research - Australia, Melbourne, VIC, Australia, \\ 
	\email{ssedai@au1.ibm.com}
}

%
%

\toctitle{Lecture Notes in Computer Science}
\tocauthor{Authors' Instructions}
\maketitle

\begin{abstract}
Optical coherence tomography (OCT) is commonly used to analyze retinal layers for assessment of ocular diseases. In this paper, we propose a  method for retinal layer segmentation and quantification of uncertainty based on Bayesian deep learning. Our method not only performs end-to-end segmentation of retinal layers, but also gives the pixel wise uncertainty measure of the segmentation output. The generated uncertainty map can be used to identify erroneously segmented image regions which is useful in downstream analysis. We have validated our method on a dataset of 1487 images obtained from 15 subjects (OCT volumes) and compared it against the state-of-the-art segmentation algorithms that does not take uncertainty into account. The proposed uncertainty based segmentation method results in comparable or improved performance, and most importantly is more robust against noise. 

\end{abstract}

\section{Introduction}
\label{sec:intro}

Optical Coherence Tomography (OCT) is a popular non-invasive imaging modality for retinal imaging. OCT provides volumetric scans of retinal layers for the diagnosis and evaluation of different diseases such as Glaucoma and Age regated macular degeneration (AMD). For example, \cite{Acton2012} have shown the correlation between outer retinal layer thickness and visual acuity in early AMD patients. It has also been shown that retinal layer features can be used to predict  vision loss and progression \cite{Farsiu2014}.

The segmentation of retinal layers in OCT has been tackled in a number of ways, such as dynamic programming \cite{Mishra2009}, graph-based shortest path algorithms\cite{Chiu2010}, graph-based minimum $s$-$t$ cut formulations \cite{Garvin2009} and level sets \cite{Carass2014,Novosel2015}.  Machine-learning based approaches have also been proposed, where the retinal layer and boundary probability maps are detected using a trained classifier. The final segmentation is then obtained by imposing a model  such as active contours \cite{Vermeer2011} or minimum $s$-$t$ cut framework \cite{Lang2013} on the soft labels.

In the past few years, Convolutional Neural Networks (CNNs) based methods such as Unet \cite{Ronneberger2015}, \cite{Sedai2017} and fully convolutional Densenet (FC-DN) \cite{Simon2017} have achieved remarkable performance gain in medical image and natural image segmentation. The networks are trained end-to-end, pixels-to-pixels on semantic segmentation exceeded the most state-of-the-art methods without further machinery. \cite{Roy2017} and \cite{Apostolopoulos2017} used Unet like network to perform pixelwise semantic segmentation of retinal layers. In another approach, \cite{Fang2017}, used CNN and graph search method for layer boundary classification.  Once trained, these methods acts as a black box where one has to assume that the segmentation output is accurate which is not always the case. For example, the model will produce incorrect segmentation when the test image is different from the distribution of images used to train the model. This may happen when the model is trained using limited number of training images. In other scenario, the model will produce inaccurate segmentation when trained using normal images, yet pathologies are observed in the test image or the test image is noisy. Quantification of uncertainties associated with the segmentation output is therefore important to determine the region of incorrect segmentation,  e.g., region associated with higher uncertainty can either be  excluded from subsequent analysis or highlighted for manual attention.  In another scenario, when the retinal layer segmentation map is used to diagnose the diseases such as AMD and Glaucoma,  the uncertainty map can be used to determine the confidence of final automatic or clinical diagnosis.

Previous works have explored the uncertainty quantification in biomedical segmentation \cite{Iglesias2013}, however, these approaches do not utilize the representative power of deep learning. Recent research has shown that Bayesian probability theory offers a mathematically grounded technique to reason about uncertainty in deep learning models\cite{Gal2016,Kandall2017}.  In this paper, we explore Bayesian fully convolutional neural network  for segmentation and uncertainty quantification of retinal layers in OCT images.   We experimentally demonstrate that in addition to the uncertainty based confidence measure, our method provides improved layer segmentation accuracy and robustness towards noise in the test images.

\section{Methodology} 
We model two types of uncertainties for retinal layer segmentation; epistemic uncertainty and aleatoric uncertainty. The epistemic uncertainty captures the uncertainty related to the model parameters, e.g., when the model does not take into account certain aspect of the training data. Therefore, the epistemic uncertainty can be reduced by training the model using more images.  Aleoretic uncertainty, on the other hand, captures the noise inherent in the images, therefore, it cannot be reduced with additional training images. We model the aleatoric uncertainty as an additional output variance for both deep learning networks. 

We enhance    fully convolutional Densenet (FC-DN) \cite{Simon2017} for  segmentation and uncertainty quantification of retinal layers.    FC-DN  is a fully convolutional neural network  with several dense-blocks connected in encoder-decoder architecture with  skip connections across them which effectively combines coarse semantic features with fine image details for pixel-wise semantic segmentation.   Each layer in the \emph{dense block} is connected to all the preceding layers by iterative concatenation of previous \emph{feature maps}. This allows all layers to access \emph{feature maps} from their preceding layers which encourages heavy feature reuse. As a result, FC-DN uses less parameter and is less prone to over-fitting. The  networks is then trained using the proposed class weighted Bayesian loss function by taking  into account the output variance which is described in Section  \ref{aleatoric_uncertainty}. Once the networks are trained, in the test phase, we use \emph{dropout variational inference} technique \cite{Gal2016} to compute the epistemic uncertainty which we describe in Section \ref{epistemic_uncertainty}

Let  $F_{\mathbf{W}}(X)$ be a  FC-DN model parameterized by $\mathbf{W}$ which takes input image $X$ and produces the \emph{logit} vector $\mathbf{z}$ for each pixel as $\mathbf{z}=F_{\mathbf{W}}(X)$. The logit vector $\mathbf{z}$ consists of \emph{logits} for each class as $\mathbf{z}=\left(z_{1,\cdots}z_{C}\right)$ where $C$ is the number of  classes i.e., number of retinal layers for segmentation.   The final probability vector for a pixel $\mathbf{y}=\left(y_{1,\cdots}y_{C}\right)$ can be computed by applying the  \emph{softmax} function over the  \emph{logits} as $\mathbf{y}=\text{Softmax (\textbf{z})}$. The \emph{softmax} function gives  the relative probabilities between classes, but fails to measure  the model's uncertainty.  

\subsection{Bayesian Fully Convolution Network}
\label{aleatoric_uncertainty}

Here we present a method to convert  FC-DN  to output the pixelwise uncertainty map in addition to the pixel-wise  segmentation map.   We name the proposed method  Bayesian FC-DN (BFC-DN).  In BFC-DN, we apply $1 \text{x}1$ convolution to the feature maps of last layers followed by \emph{softplus} activation to output the   variance $\mathbf{v}$ for each pixel in addition to the \emph{logit} vector $\mathbf{z}$ i.e.,  $(\mathbf{z},\mathbf{v})=F_{\mathbf{W}}(X)$.  This variance  gives aleatoric uncertainty of the model  which the  network learns to predict  during the training.  In addition, we include the \emph{dropout layer}  before every convolution layer which allows us to compute \emph {epistemic uncertainty} which will be described in Section \ref{epistemic_uncertainty}.

 The output of the model is the Gaussian distribution $\mathcal{{N}}\left(\mathbf{z},\mathbf{v}\right)$. Computing  the categorical cross entropy loss over this distribution    is not feasible. Therefore, we approximated it using the monte-carlo integration. Given a set of training images and corresponding ground truth segmentation mask, $\ensuremath{D=\left\{ X_{n},Y_{n}\right\} _{n=1}^{N}}$, output \emph{logit} for each sample in the mini-batch is perturbed $T$  times with a Gaussian noise $\epsilon_{t}\sim\mathcal{{N}}\left(0,\mathbf{v}\right)$ as $\hat{ \mathbf{z_{t}}}=\textbf{z}+\epsilon_{t}$ and the final pixel-wise  bayesian loss  is computed as:
 
\begin{equation}
L(W)=-\frac{1}{T}\sum_{t=1}^{T}\sum_{c=1}^{C}\beta_{c}\sum_{ \forall Y_{c}}\log y_{c}^{t} \label{bayesian_loss}
\end{equation}
where $y_{c}^{t}$ is obtained by applying \emph{softmax} to the \emph{logit} vector $\hat{ \mathbf{z_{t}}}$;  $Y_{c}$ denotes the pixels region of the  $c^{th}$ class in the ground truth Y and the scale factor $\beta_{c}=1/|Y_{c}|$ weights the contribution of each class to mitigate the class imbalances of different OCT layers and the background by increasing the weight of under represented classes  while decreasing the effect of over represented classes. The proposed Bayesian loss function encourage the network to minimize the larger losses by increasing the variance, therefore is more robust towards noise..

We  train the proposed  BFC-DN using the \emph{bayesian loss}  given by Equation \ref{bayesian_loss} for 40000 iterations. We have used mini-batch gradient descent and the Adam optimizer with momentum and a batch size of 2. The learning rate is set to $10^{-5}$ which is decreased by one tenth after 10000 iterations of the training. Data augmentation is an important step in training deep networks. We augment the training images and corresponding label map masks through a mirror-image reflection and random rotation within the range of $[-15, 15]$ degrees.

\subsection{Segmentation and Uncertainty Quantification}
\label{epistemic_uncertainty}

Epistemic uncertainty  is generally computed by assuming  distribution over the network weights which allows  the computation of  distribution of class probabilities rather than point estimate \cite{NIPS2016_6117}. Such methods require   optimization over weights distribution and therefore is computationally expensive \cite{Gal2016}.   We adopt more practical approach introduced by \cite{Gal2016} which is based on the \emph{dropout variational inference}.  We  train the  BFC-DN  with a \emph{dropout} layer before every convolution layer and use the \emph{dropout} in test phase as well.  Specifically, segmentation samples from the output predictive distribution are obtained by performing $T$   stochastic forward passes through the network, i.e.,  $(\mathbf{z}^{t},\mathbf{v}^{t})=F_{\mathbf{\hat{W}}_{t}}(X), t=1,\cdots, T$ where $\mathbf{\hat{W}}_{t}$  is an effective network weight  after the  \emph{dropout}. In each forward pass, the fraction of network weights (denoted by \emph{dropout rate}) are disabled and the segmentation score is computed using only the remaining weights. The segmentation score  vector $\mathbf{\bar{y}}$ and the  aleatoric variance $\mathbf{\bar{v}}$  is obtained by averaging the $T$ samples, via \emph{monte carlo integration}:
\begin{eqnarray}
\mathbf{\bar{y}} & = & \frac{1}{T}\sum_{t=1}^{T}\text{Softmax}(\mathbf{z}^{t})  \label{segmentation_output}\\
\mathbf{\bar{v}} & = & \frac{1}{T}\sum_{t=1}^{T}\mathbf{v}^{t}  \label{average_alereotic}
\end{eqnarray}

The average score vector contains the probability score  for each retinal layers class, ie $\mathbf{\bar{y}} = [\bar{y}_1,\cdots,\bar{y}_C ] $. The overall segmentation uncertainty for each pixel can then be obtained as:

\begin{equation}
U(\mathbf{\bar{y}})=-\sum_{c=1}^{C}\bar{y}_{c}\text{log }\bar{y}_{c} \label{entropy} +\mathbf{\bar{v}}
\end{equation}

where the first term  denotes  \emph{epistemic uncertainty}  of the score computed as the  entropy of the average score vector obtained by averaging $T$ stochastic predictions (Equation \ref{segmentation_output}) and the second term is the uncertainty output produced by the network itself (Equation \ref{average_alereotic}). We set the \emph{dropout} rate=0.4 and  $T=50$ to allow sufficient sampling of network weights for final prediction.

For uncertain predictions, network assigns higher probabilities to different classes for different forward passes, resulting in higher epistemic uncertainty  given by Equation \ref{entropy}. For the certain predictions, network assigns higher probability to the true class for different forward passes, resulting in lower epistemic uncertainty.  Since epistemic uncertainty is  related to the model parameters weights, it can be reduced by observing more data. This is because, the network becomes robust towards  weight dropout in test phase as it observes  more data.

\section{Experiments}

The dataset \cite{Srinivasan:14}  consists of $1487$  images from $15$ spectral-domain optical coherence tomography (OCT) volumes from unique normal subjects acquired  on a Spectralis scanner. The size of each volume is $512  \times 496 \times N_{slices} $ where $N_{slices}$ is different for each volume and ranges from $49-100$. All scans have axial resolution of $3.87\upmu$m. The ground truth has been obtained by manual annotation  of the nine boundaries  from eight retinal layers \cite{Lang2013}. To facilitate the pixel-wise semantic segmentation, we convert the layer boundaries to the probability map for the eight layers regions and the background region. Therefore, the number of classes is $C=9$.

Out of $1487$ images, we select $1116$ images from $12$ volumes to create a training  set and remaining $291$ images from $3$ volumes for validation. We compare our method with the baseline  FC-DN \cite{Simon2017} which do not take into account uncertainty, i.e, the networks do not output aleatoric variance and segmentation is performed in a single forward pass by disabling the  \emph{dropout} is the test phase. To train these networks, we use non-bayesian class weighted cross entropy loss function which can be derived by setting $T=1$ and $v=0$ in Equation \ref{bayesian_loss}. 

\begin{table}[ht]
	\caption{Performance of our proposed retinal layer segmentation method compared
		with the state-of-the-art Jegou  et. el. \cite{Simon2017} and Lang
		et. el. \cite{Lang2013} segmentation methods.}
	\label{tab:res}
	\centering{}%
	\begin{tabular}{|c|c|c|c|c|c|}
		\hline 
		Layer & \multicolumn{2}{c|}{Dice Coefficient} & Boundary & \multicolumn{2}{c|}{Absolute error $\upmu$m}\tabularnewline
		\hline 
		& FC-DN \cite{Simon2017} & \textbf{BFC-DN} &  &  Lang et. el.\cite{Lang2013}  & \textbf{BFC-DN}\tabularnewline
		\hline 
		RNFL & $0.94\pm0.01$ & $0.95\pm0.01$ & ILM & $2.6\pm3.89$ & $1.81\pm4.12$\tabularnewline
		\hline 
		GCL+IPL & $0.96\pm0.01$ & $0.97\pm0.01$ & RNFL-GCL & $4.0\pm6.11$ & $3.6\pm6.3$\tabularnewline
		\hline 
		INL & $0.91\pm0.01$ & $0.93\pm0.01$ & IPL-INL & $3.78\pm4.41$ & $2.6\pm2.73$\tabularnewline
		\hline 
		OPL & $0.90\pm0.01$ & $0.91\pm0.01$ & INL-OPL & $3.66\pm3.84$ & $2.9\pm2.8$\tabularnewline
		\hline 
		ONL & $0.96\pm0.008$ & $0.96\pm0.005$ & OPL-ONL & $3.4\pm4.24$ & $2.64\pm2.54$\tabularnewline
		\hline 
		IS & $90\pm0.01$ & $0.91\pm0.01$ & ELM & $2.79\pm2.68$ & $2.44\pm2.4$\tabularnewline
		\hline 
		OS & $0.91\pm0.01$ & $0.92\pm0.01$ & IS-OS & $2.38\pm2.49$ & $2.1\pm2.3$\tabularnewline
		\hline 
		RPE & $0.95\pm0.01$ & $0.96\pm0.008$  & OS-RPE & $4.16\pm4.13$ & $3.8\pm3.3$\tabularnewline
		\hline 
		- & - & - & BrM & $3.87\pm3.69$ & $3.14\pm2.81$\tabularnewline
		\hline 
	\end{tabular} 
\end{table}

Table \ref{tab:res} compares the average Dice coefficient (DC) between the ground truth and predicted segmentation of the $8$ layers using the proposed Bayesian method ( BFC-DN) and non-Bayesian method (FC-DN \cite{Simon2017}). The proposed  method BFC-DN resulted in highest DC of $0.97$ for GCL+IPL layer and lowest DC of  $0.91$ for OPL and IS layer. Moreover, BFC-DN  resulted in improved segmentation  for most of the layers in comparison to FC-DN.    Table \ref{tab:res} also compares the average absolute error   for $9$ boundaries  of our method   with  \cite{Lang2013}. We observe that  BFC-DN    resulted in lower  error than  \cite{Lang2013}    which indicates proposed uncertainty based method is  effective in segmenting retinal layers.  

\begin{figure}[t]
	
	\centering \includegraphics[scale=0.2]{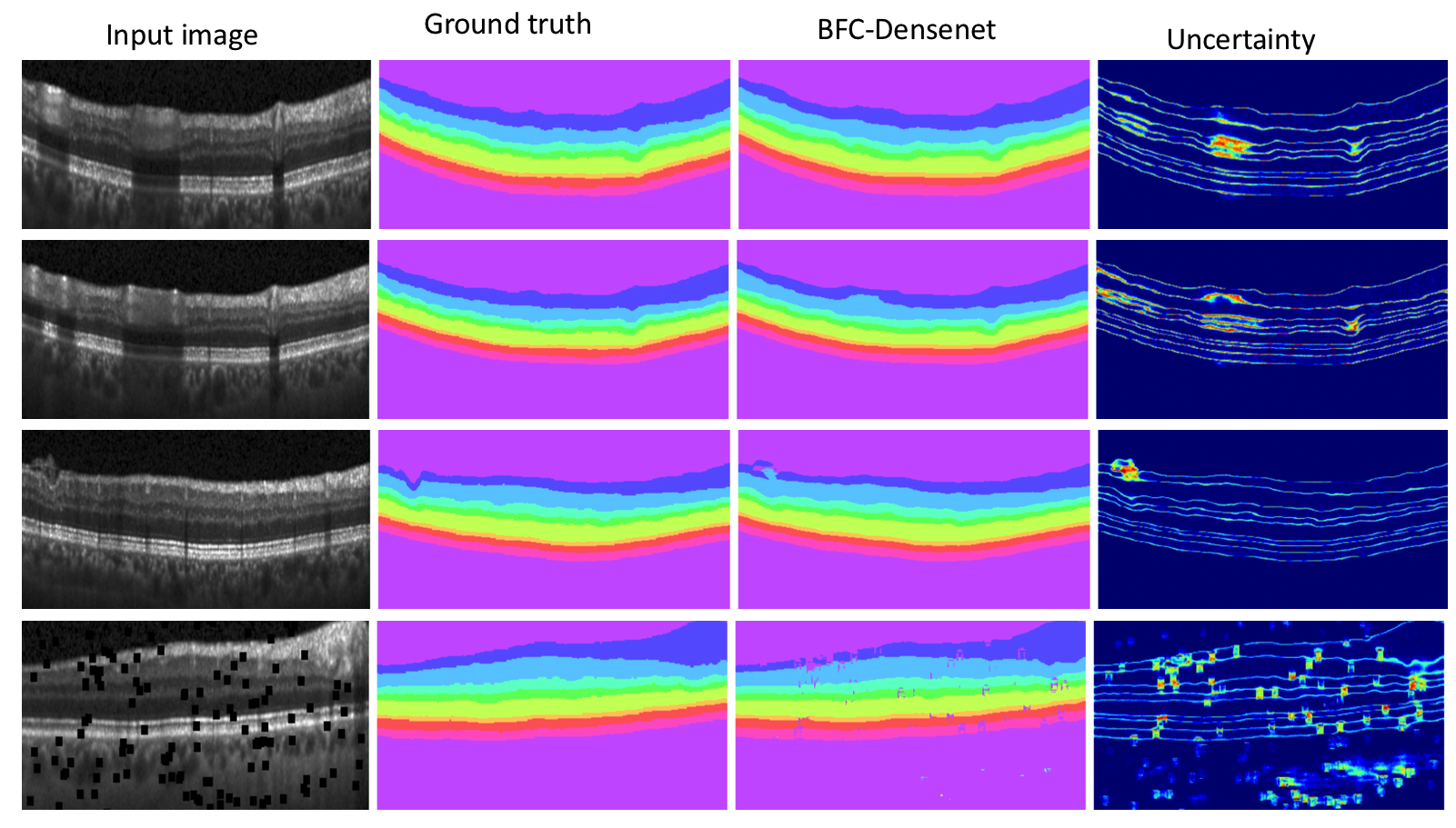}
	\caption{Examples of retinal layer segmentation and uncertainty quantification using proposed BFCN-Densenet. (a) test images, (b) ground truth, (c) predicted segmentation map from FBC-Densenet  (d)   uncertainty map  (warmer color denotes regions with higher uncertainty). The last row shows  an example of layer segmentation in test images with added random block noise.}	\label{result1}
\end{figure}

\begin{figure}
	\centering \includegraphics[scale=0.25]{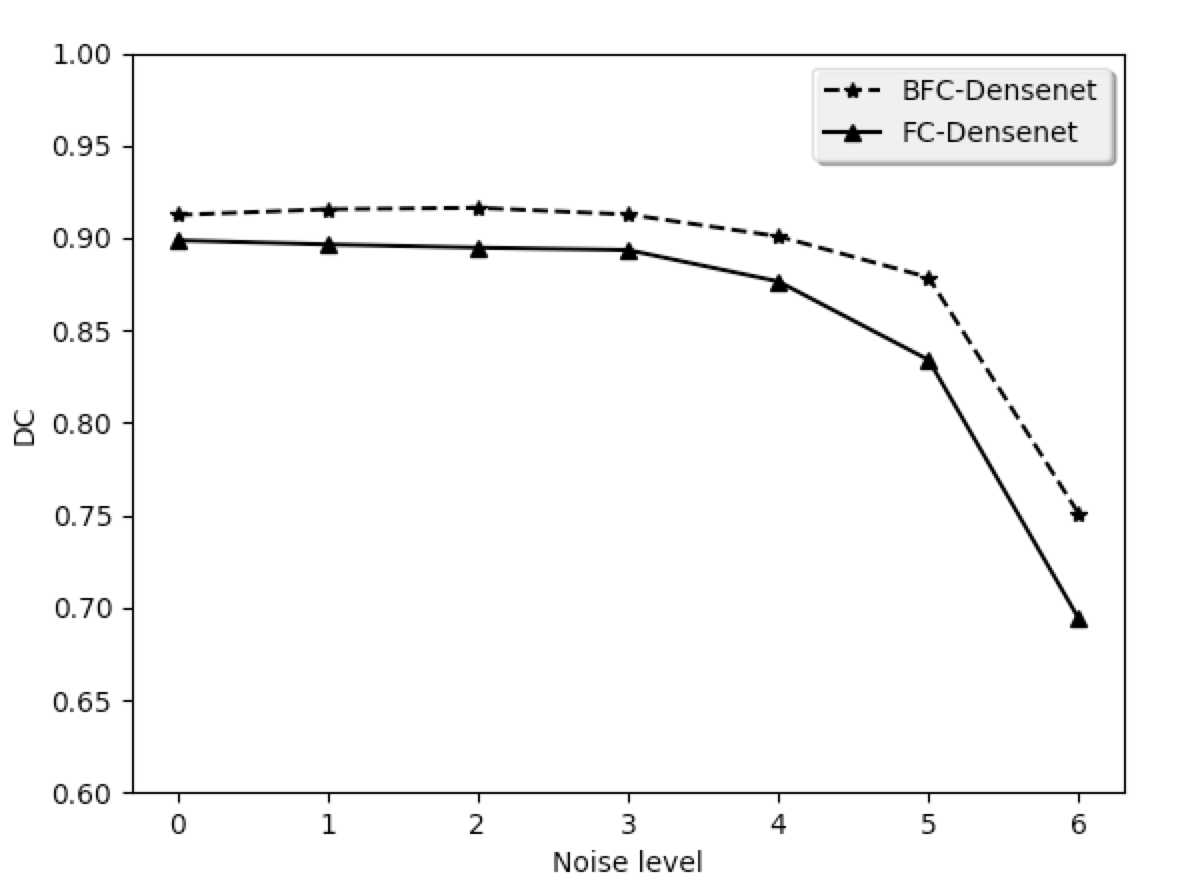} 
	\caption{Comparison of average segmentation performance of proposed BFC-DN with FC-DN \cite{Simon2017} for different noise levels. The number of  random block noise components at a given noise level is double than that  at previous  level. }  \label{fig:result_noise}
	
\end{figure}

Figure \ref{result1} shows the examples of segmentation and uncertainty maps produced by our proposed method on few images from the validation set. It can be seen that our method produces pixel-wise uncertainty associated with the segmentation output where high  uncertainty  correlates with the inaccurate segmentation in the corresponding  region.  In order to validate the robustness of our proposed method against noise, we evaluate the performance by adding random block noise to the test images  as  shown in the last row image of  Figure \ref{result1}. We observe that  BFC-DN performs much better than FC-DN in presence of large noise levels as shown in Figure \ref{fig:result_noise}. This demonstrates that BFC-DN is more robust towards the noisy images than FC-DN.

The average execution time for the retinal layer segmentation  for BFC-DN is 2.5 seconds per image Tesla-K40 GPU which is somewhat slower than that of FC-DN which took  300 ms.   This is because our model requires $T$ forward passes in the test phase in contrast to FC-DN which requires one forward pass.

\section{Conclusion} 
In this paper, we proposed a Bayesian deep learning based method for retinal layer segmentation in OCT images. Our method produces layer segmentation and corresponding uncertainty maps depicting the pixel-wise confidence measure of the segmentation output. Experimental results demonstrate that our method compares favorably with non-bayesian DL methods, particularly in the presence of noise and outperforms sate of the art boundary based segmentation method. We have shown qualitatively that the resulting uncertainty maps correlates with the inaccuracies in segmentation output. The proposed method is applicable in determining the confidence of image analysis modules that utilizes the segmentation output for downstream analysis. Such uncertainty visualization can also be useful in computer-assisted diagnostic systems where clinician have additional insight about various measurements generated by the system to make necessarily adjustments and make more informed decisions Also, the resulting uncertainty map can be integrated within active learning systems to correct the segmentation output.

\bibliographystyle{splncs03}
\bibliography{refs}

\begin{thebibliography}{10}
\providecommand{\url}[1]{\texttt{#1}}
\providecommand{\urlprefix}{URL }

\bibitem{Acton2012}
Acton, J.H., Smith, R.T., Hood, D.C., , Greenstein, V.C.: Relationship between
  retinal layer thickness and the visual field in early age-related macular
  degeneration. Investigative Ophthalmology and Visual Science  53(12),
  7618--7624 (2012)

\bibitem{Apostolopoulos2017}
Apostolopoulos, S., Zanet, S.D., Ciller, C., Wolf, S., Sznitman, R.:
  Pathological {OCT} retinal layer segmentation using branch residual u-shape
  networks. CoRR  abs/1707.04931 (2017)

\bibitem{Carass2014}
Carass, A., Lang, A., Hauser, M., Calabresi, P.A., Ying, H.S., Prince, J.L.:
  {Multiple-object geometric deformable model for segmentation of macular OCT}.
  Biomedical Optics Express  5(4),  1062 (2014)

\bibitem{Chiu2010}
Chiu, S.J., Li, X.T., Nicholas, P., Toth, C.A., Izatt, J.A., Farsiu, S.:
  {Automatic segmentation of seven retinal layers in SDOCT images congruent
  with expert manual segmentation}. Optics express  18(18),  19413--28 (2010)

\bibitem{Fang2017}
Fang, L., Cunefare, D., Wang, C., Guymer, R.H., Li, S., Farsiu, S.: Automatic
  segmentation of nine retinal layer boundaries in oct images of non-exudative
  amd patients using deep learning and graph search. Biomedical Optics Express
  8(5),  2732 (2017)

\bibitem{Farsiu2014}
Farsiu, S., J.Chiu, S., V.O'Connell, R., A.Folgar, F., Yuan, E., A.Izatt, J.,
  A.Toth, C.: Quantitative classification of eyes with and without intermediate
  age-related macular degeneration using optical coherence tomography.
  ophthalmology. Opthamalogy  121(1),  162--72 (2014)

\bibitem{Gal2016}
Gal, Y., Ghahramani, Z.: Dropout as a bayesian approximation: Representing
  model uncertainty in deep learning. In: International Conference on
  International Conference on Machine Learning. pp. 1050--1059 (2016)

\bibitem{Garvin2009}
Garvin, M.K., Abr{\`{a}}moff, M.D., Wu, X., Russell, S.R., Burns, T.L., Sonka,
  M.: {Automated 3D intraretinal layer segmentation of macular spectral-domain
  optical coherence tomography images}. IEEE Transactions on Medical Imaging
  28(9),  1436--1447 (2009)

\bibitem{Iglesias2013}
Iglesias, J.E., Sabuncu, M.R., Leemput, K.V.: Improved inference in bayesian
  segmentation using monte carlo sampling: Application to hippocampal subfield
  volumetry. Medical Image Analysis  17(7),  766--778 (2013)

\bibitem{Simon2017}
J{\'{e}}gou, S., Drozdzal, M., V{\'{a}}zquez, D., Romero, A., Bengio, Y.: The
  one hundred layers tiramisu: Fully convolutional densenets for semantic
  segmentation. In: 2017 {IEEE} Conference on Computer Vision and Pattern
  Recognition Workshops, {CVPR} Workshops, Honolulu, HI, USA, July 21-26, 2017.
  pp. 1175--1183 (2017)

\bibitem{Kandall2017}
Kendall, A., Gal, Y.: {What Uncertainties Do We Need in Bayesian Deep Learning
  for Computer Vision?} In: Neural Information Processing Systems (NIPS) (2017)

\bibitem{Lang2013}
Lang, A., Carass, A., Hauser, M., Sotirchos, E.S., Calabresi, P.A., Ying, H.S.,
  Prince, J.L.: {Retinal layer segmentation of macular OCT images using
  boundary classification}. Biomedical optics express  4(7),  1133--1152 (2013)

\bibitem{Mishra2009}
Mishra, A., Wong, A., Bizheva, K., Clausi, D.A.: {Intra-retinal layer
  segmentation in optical coherence tomography images.} Optics Express  17(26),
   23719--28 (2009)

\bibitem{Novosel2015}
Novosel, J., Thepass, G., Lemij, H.G., de~Boer, J.F., Vermeer, K.A., van Vliet,
  L.J.: {Loosely coupled level sets for simultaneous 3D retinal layer
  segmentation in optical coherence tomography}. Medical Image Analysis  26(1),
   146--158 (2015)

\bibitem{Ronneberger2015}
Ronneberger, O., Fischer, P., Brox, T.: U-Net: Convolutional Networks for
  Biomedical Image Segmentation, pp. 234--241. Springer, Cham (2015)

\bibitem{Roy2017}
Roy, A.G., Conjeti, S., Karri, S.P.K., Sheet, D., Katouzian, A., Wachinger, C.,
  Navab, N.: Relaynet: Retinal layer and fluid segmentation of macular optical
  coherence tomography using fully convolutional network. CoRR  abs/1704.02161
  (2017)

\bibitem{Sedai2017}
Sedai, S., Tennakoon, R., Roy, P., Cao, K., Garnavi, R.: Multi-stage
  segmentation of the fovea in retinal fundus images using fully convolutional
  neural networks. In: ISBI. pp. 1083--1086 (April 2017)

\bibitem{NIPS2016_6117}
Springenberg, J.T., Klein, A., Falkner, S., Hutter, F.: Bayesian optimization
  with robust bayesian neural networks. In: Lee, D.D., Sugiyama, M., Luxburg,
  U.V., Guyon, I., Garnett, R. (eds.) Advances in Neural Information Processing
  Systems 29, pp. 4134--4142. Curran Associates, Inc. (2016)

\bibitem{Srinivasan:14}
Srinivasan, P.P., Kim, L.A., Mettu, P.S., Cousins, S.W., Comer, G.M., Izatt,
  J.A., Farsiu, S.: Fully automated detection of diabetic macular edema and dry
  age-related macular degeneration from optical coherence tomography images.
  Biomed. Opt. Express  5(10),  3568--3577 (Oct 2014)

\bibitem{Vermeer2011}
Vermeer, K.A., van~der Schoot, J., Lemij, H.G., de~Boer, J.F.: {Automated
  segmentation by pixel classification of retinal layers in ophthalmic OCT
  images.} Biomedical optics express  2(6),  1743--1756 (2011)

\end{thebibliography}

\end{document}